\documentclass{article}

\usepackage{microtype}
\usepackage{graphicx}
\usepackage{subcaption}
\usepackage[section]{placeins}
\usepackage{booktabs} 
\usepackage{amssymb}
\usepackage{adjustbox}
\usepackage{balance}
\usepackage{tabularx}

\usepackage{hyperref}

\usepackage{amsmath}

\usepackage[accepted]{icml2019}

\icmltitlerunning{Pretraining \& Reinforcement Learning}

\begin{document}

\twocolumn[
\icmltitle{Pretraining \& Reinforcement Learning:\\Sharpening the Axe Before Cutting the Tree 
}



\icmlsetsymbol{equal}{*}

\begin{icmlauthorlist}
\icmlauthor{Saurav Kadavath}{equal,to}
\icmlauthor{Samuel Paradis}{equal,to}
\icmlauthor{Brian Yao}{equal,to}
\end{icmlauthorlist}

\icmlaffiliation{to}{University of California, Berkeley, USA}


\icmlkeywords{Machine Learning, ICML}

\vskip 0.3in
]



\printAffiliationsAndNotice{\icmlEqualContribution} 

\begin{abstract}
Pretraining is a common technique in deep learning for increasing performance and reducing training time, with promising experimental results in deep reinforcement learning (RL). However, pretraining requires a relevant dataset for training. In this work, we evaluate the effectiveness of pretraining for RL tasks, with and without distracting backgrounds, using both large, publicly available datasets with minimal relevance, as well as case-by-case generated datasets labeled via self-supervision. Results suggest filters learned during training on less relevant datasets render pretraining ineffective, while filters learned during training on the in-distribution datasets reliably reduce RL training time and improve performance after 80k RL training steps. We further investigate, given a limited number of environment steps, how to optimally divide the available steps into pretraining and RL training to maximize RL performance. Our code is available on GitHub \footnote{\url{github.com/ssss1029/DeepRL_Pretraining}}.\parfillskip=0pt
\end{abstract}


\section{Introduction}
Pretraining, or using model parameters from a previous task to initialize model parameters for a new task, is common in deep learning. Pretraining often increases performance, improves robustness, and reduces training time. As a result, the technique naturally lends itself to deep reinforcement learning, which often requires a prohibitively large amount of training time and data to reach reasonable performance. For example, work such as \citet{cruz2018pretraining} and \citet{rajapakshe2020deep} use pretraining in an attempt to separate the learning of the feature representations from the learning of the policy. Both demonstrate that using pretraining improves both training speed and convergence score in a variety of RL settings. However, pretraining requires a relevant dataset for training. \citet{rajapakshe2020deep} use a large, public dataset that is relevant to their goal task. \citet{cruz2018pretraining} and \citet{cruz2019pretraining} collect labeled datasets from human demonstrations on the target task in order to pretrain the network.

In this work, we explore methods for pretraining networks for RL in the case where a relevant dataset is not publicly available or collecting human demonstrations is infeasible. We pretrain convolutional neural networks on image-based tasks, and use the convolutional weights to initialize a network for RL training. Similar to previous work, the goal is to separate the learning of feature representations from the learning of the policy. By learning useful features during pretraining, the network is able to focus on the policy during RL training. We evaluate the performance of these methods on control tasks from the DeepMind Control Suite using Soft Actor-Critic \citep{tassa2018deepmind, haarnoja2018soft}.

For general purpose pretraining, we evaluate the transferability of networks for pixel-based RL agents that are pretrained using large, well-annotated datasets with no relevance to the RL task, such as ImageNet. Despite the distribution mismatch, we hypothesize low-level filters may still be useful for feature extraction in RL. However, results suggest the effectiveness of ImageNet pretraining is limited on the DeepMind Control Suite.

We also explore an additional pretraining strategy based on learning the inverse kinematics (IK) of each individual environment. Using self-supervision, we generate a labeled dataset $\mathcal{D}_{\rm env} = \left\{(o_t, o_{t+1}, a_t)\right\}_{t=1}^{T}$, where $o_{t+1}$ is the observation that results from taking action $a_t$ from observation $o_t$. We then use $\mathcal{D}_{\rm env}$ to pretrain a network via supervised learning. Through self-supervision, we maintain the generalizability of the method while reducing the distribution mismatch between the pretraining data and the RL observations. Although this method of pretraining requires access to the environment beforehand, it does not require any knowledge of the reward structure, lending itself useful for complex real-world tasks where an unbiased reward is difficult to programmatically quantify. We evaluate the performance of RL agents initialized with weights from IK pretraining, with results suggesting this method is able to reduce RL training time and improve performance after 80k RL training steps for all environments. We also evaluate in the effectiveness of IK pretraining when the pretraining environment is different from the RL training environment. In order to evaluate this, we pretrain and test on all pairs of different environments. We find cross-environment pretraining substantially helps RL performance for some environments, while it hurts in others. 

We further evaluate the above experiments on RL tasks with distracting backgrounds. Distracting backgrounds often hinder performance on RL tasks since they convolute the feature extraction process. However, since the pretrained network extracts general features prior to RL training, we hypothesize pretraining can increase learning speed and improve performance in this setting. Distractions are added by playing a $\sim$20 second video clip in the background during pretraining and RL training. Interestingly, both pretraining methods are comparatively less helpful in the case of distracting backgrounds.

Lastly, we consider the case where environment steps are expensive, and thus need to be limited. For motivation, a famous proverb states ``give me six hours to chop down a tree, and I will spend the first five sharpening the axe." 
In our case, given a limited number of environment steps, we investigate how to optimally divide the steps into pretraining and RL training to maximize RL performance. Results across all 3 environments suggest the benefits of IK pretraining are limited when the number of environment steps allowed is small. In other words, while sharpening the axe is often useful, if you only have 5 minutes to chop, you are better off working with what you have.

\subsection{Key Contributions}
\begin{enumerate}
    \item Analysis of a general-purpose pretraining technique for deep reinforcement learning.
    \item Results suggesting a novel self-supervised pretraining technique reliably reduces RL training time and improves performance. 
    \item Analysis of optimizing RL performance in the case the total number of environment steps is limited.
\end{enumerate}

\section{Related Works}

\subsection{Pretraining in Other Domains}

Pretraining networks in vision has been shown to improve accuracy on downstream tasks. Previous works improve performance on ImageNet by pretraining a network on a massive external dataset, and then fine-tuning on ImageNet \citep{hendrycks2020faces}. Along with improving performance, the authors notice large improvements in robustness; the pretraining helped their final classifier generalize to unseen distributions significantly better than a naively trained network. Other work in computer vision also use pretraining to improve robustness to adversarial examples, label corruption, and class imbalance for image classifiers \citep{hendrycks2019using}. \citet{rebuffi2017learning} develop pretraining techniques to learn a single visual representation, and then apply it to several different tasks, such as classifying different categories of images. 

Pretraining is also used extensively in NLP to improve data efficiency for downstream tasks and improve robustness of models \citep{devlin2019bert, hendrycks2020pretrained, brown2020language}. Large language models are often pretrained in an unsupervised manner on a large corpus of text, allowing them to learn language before learning how to solve more specific problems such a Question-Answering, Token Classification, and Sentiment Detection. This greatly improves performance over any known method that does not make use of this pretraining step.

\subsection{Pretraining in RL}

Various papers have explored the use of pretraining in the reinforcement learning setting. \citet{cruz2019pretraining} applied pretraining by using human demonstrations to collect a labeled dataset mapping states to actions, training a classifier on this dataset, and then using the weights from this classifier to initialize a deep Q-network (as opposed to random initialization). By learning features in this pretraining phase, the authors are able to significantly improve training speed in a number of Atari environments, as well as improve convergence score for various tasks. In other work, the authors show that learning from non-expert, noisy human demonstrations has similar benefits to learning from expert human demonstrations \citep{cruz2018pretraining}, suggesting pretraining assists with feature extraction more than policy extraction. \citet{rajapakshe2020deep} study the use of pretraining in deep reinforcement learning for speech recognition, and find that pretraining a classifier on a public speech dataset results in higher accuracy within a reduced time than random initialization. \citet{larsson2018} conducts a comprehensive exploration on pretraining in RL, including transferring weights from the first layer of a trained CNN prior to RL training, and concludes with experimental results suggesting pretraining a network to learn features has benefits in the deep RL setting in terms of learning speed and convergence score. The heuristic that unites the above papers is separating feature extraction from policy learning via pretraining on a relevant, labeled dataset. 

\subsection{Soft Actor-Critic}
\label{subsec:sac}


\begin{figure}[t]
\begin{subfigure}{.155\textwidth}
  \centering
    \includegraphics[width=\textwidth]{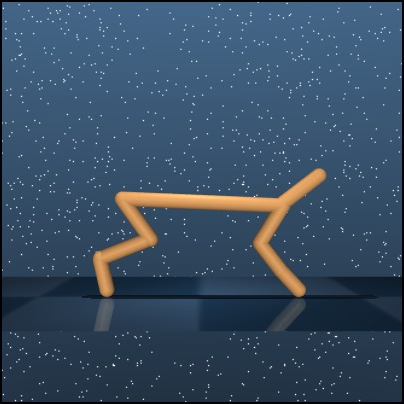}
  \caption{Cheetah}
  \label{fig:environments_a}
\end{subfigure}%
\hspace*{.25pt}
\begin{subfigure}{.155\textwidth}
  \centering
    \includegraphics[width=\textwidth]{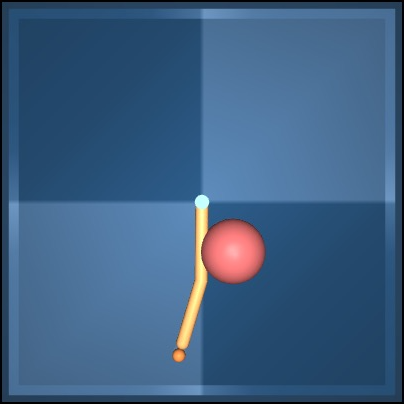}
  \caption{Reacher}
  \label{fig:environments_b}
\end{subfigure}%
\hspace*{.25pt}
\begin{subfigure}{.155\textwidth}
  \centering
    \includegraphics[width=\textwidth]{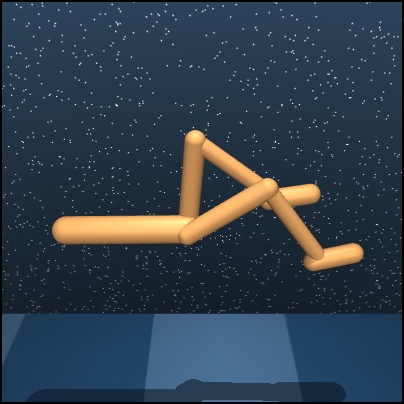}
  \caption{Walker}
  \label{fig:environments_c}
\end{subfigure}\\

\vspace*{3pt}

\begin{subfigure}{.155\textwidth}
  \centering
    \includegraphics[width=\textwidth]{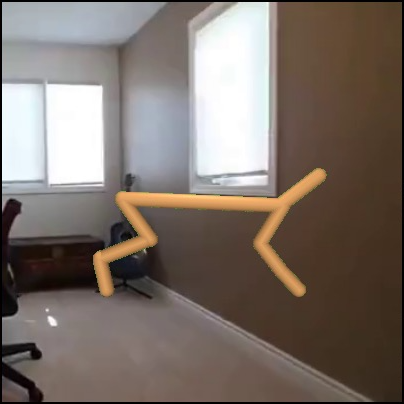}
  \caption{Distraction 1}
  \label{fig:environments_d}
\end{subfigure}%
\hspace*{.25pt}
\begin{subfigure}{.155\textwidth}
  \centering
    \includegraphics[width=\textwidth]{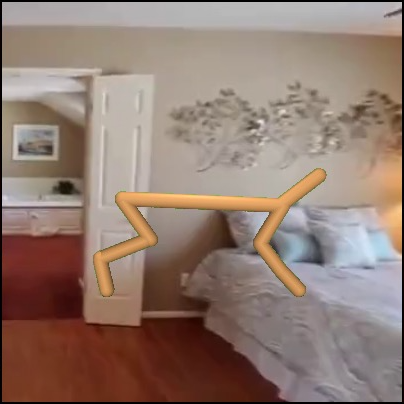}
  \caption{Distraction 2}
  \label{fig:environments_e}
\end{subfigure}%
\hspace*{.25pt}
\begin{subfigure}{.155\textwidth}
  \centering
    \includegraphics[width=\textwidth]{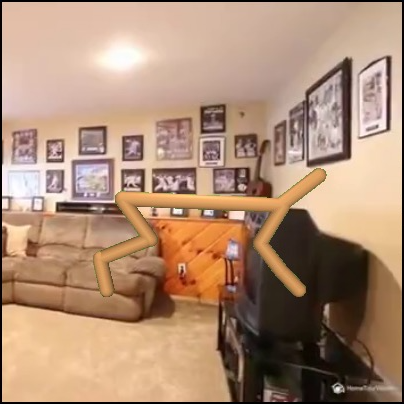}
  \caption{Distraction 3}
  \label{fig:environments_f}
\end{subfigure}%
\vspace*{-3pt}
\caption{\textbf{DM Control Tasks.} We evaluate the performance of our pretraining methods using control tasks from the DeepMind Control Suite \citep{tassa2018deepmind}. In particular, the \texttt{run} task in the Cheetah environment as shown in (a), the \texttt{easy} task in the Reacher environment as shown in (b), and the \texttt{walk} task in the Walker environment as shown in (c). We also evaluate performance of our pretraining methods using the above tasks with distracting backgrounds added. Distractions are added by playing a $\sim$20 second video clip in the background during pretraining and RL training. 3 different videos are used, shown on the Cheetah environment in (d), (e), \& (f). 
}
\label{fig:environments}
\end{figure} 

Soft Actor-Critic (SAC) is an off-policy algorithm for continuous control  \citep{haarnoja2018soft}. At its core, the goal of SAC is to simultaneously maximize expected return and entropy (a measure of policy randomness). Doing so results in an actor that is able to successfully complete a task while maximizing randomness in its actions, which helps promote exploration and avoid premature convergence to local optima. The result is a more robust, stable policy.


\section{Experimental Design} 

\subsection{Environments}

\subsubsection{DM Control}

We evaluate the performance of our pretraining methods using control tasks from the DeepMind Control Suite \citep{tassa2018deepmind}. In particular, we test on three tasks. First, the \texttt{run} task in the Cheetah environment, in which a bipedal agent earns a reward for moving with high velocity (see Figure~\ref{fig:environments_a}). Second, the \texttt{easy} task in the Reacher environment, where the agent aims to contact a sphere with its end effector (see Figure~\ref{fig:environments_b}). Third, the \texttt{walk} task in the Walker environment, where the goal of the agent is to stand upright and obtain forward velocity (see Figure~\ref{fig:environments_c}).

\subsubsection{Distracting Environments}

We also evaluate the performance of our pretraining methods using the above tasks with distracting backgrounds added. We add distractions via a $\sim$20 second video playing in the background. Distracting backgrounds often hinder performance on RL tasks, since it convolutes the feature extraction process. However, since the pretrained network extracts general features prior to RL training, we hypothesize pretraining can increase learning speed and improve performance. Figure~\ref{fig:environments_d}-\ref{fig:environments_f} illustrate a random frame from each of the 3 unique background videos on the Cheetah environment.

\subsection{General Purpose Pretraining: ImageNet}

\begin{figure}[t!]
  \centering
    \includegraphics[width=.47\textwidth]{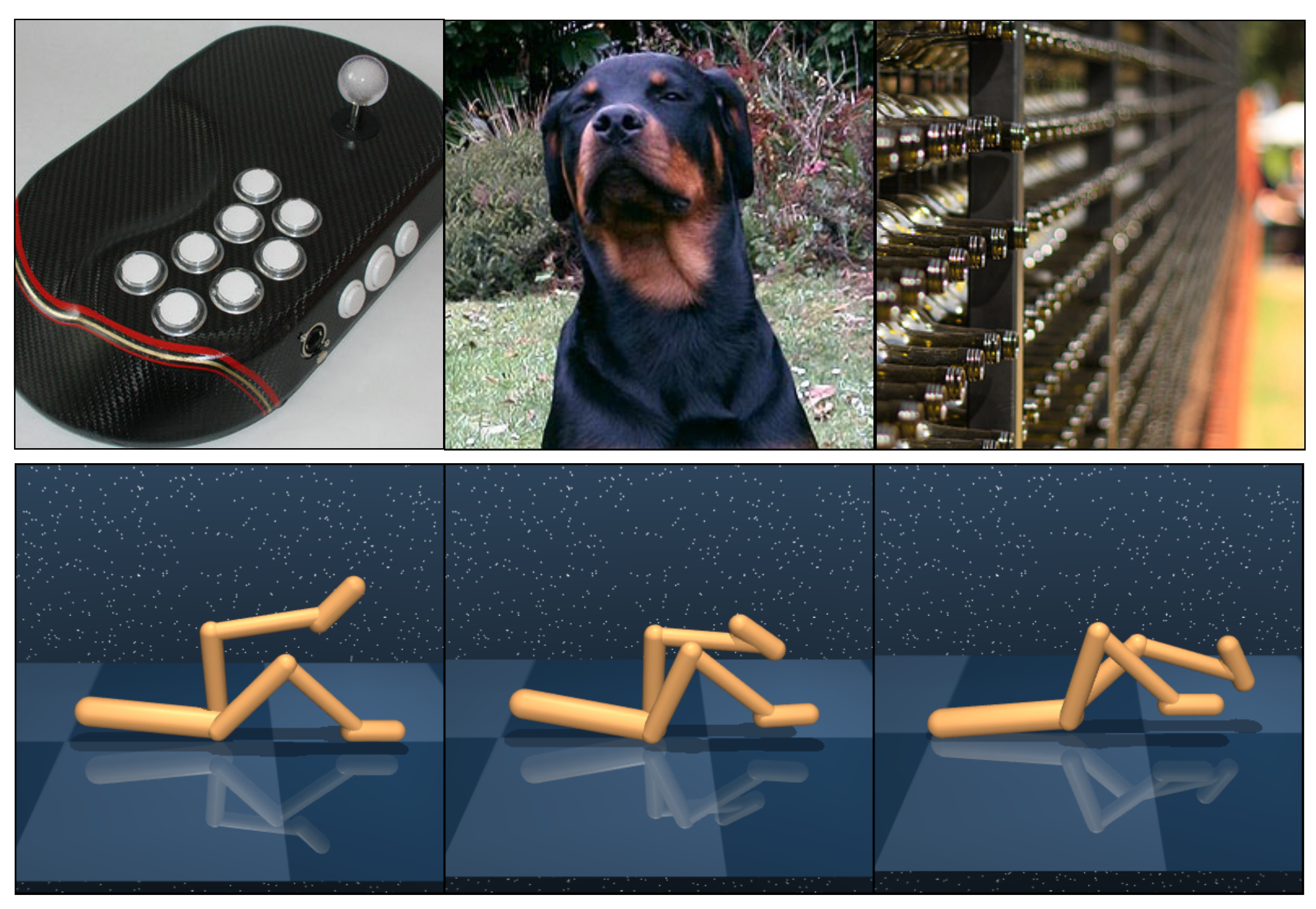}
\vspace{-8pt}
\caption{\textbf{ImageNet Samples vs. Walker Observations.} A clear distribution mismatch exists between ImageNet and DM Control. However, we hypothesize that the ImageNet classification task is in some sense harder due to diversity, and starting with the low-level filters that are learned using ImageNet training will still be more helpful than starting with a randomly initialized network for RL training.
}
\label{fig:imagenet_walker_comparison}
\end{figure}

Our first method of pretraining networks for pixel-based RL agents involves leveraging large, well-annotated datasets that are commonly used by the supervised learning community. Our goal is to learn weights for a convolutional feature extractor by training on ImageNet's supervised classification task, and then loading these weights from our pretrained feature extractor at the beginning of RL training. This should allow RL agents to spend less time learning how to do feature extraction and more time learning good policies.

We note a distributional mismatch exists between ImageNet samples and observations from the DM Control environments, as shown in Figure~\ref{fig:imagenet_walker_comparison}. However, we hypothesize that a network pretrained on ImageNet will still learn low-level filters that are useful for RL, since the subtasks of detecting lines and shapes should remain similar across distributions. We later experiment with a different pretraining task, in which the pretraining samples more closely match observations seen during our RL tasks.

\subsubsection{Architecture \& Pretraining Setup}

\begin{figure}[t]
  \centering
    \includegraphics[width=.47\textwidth]{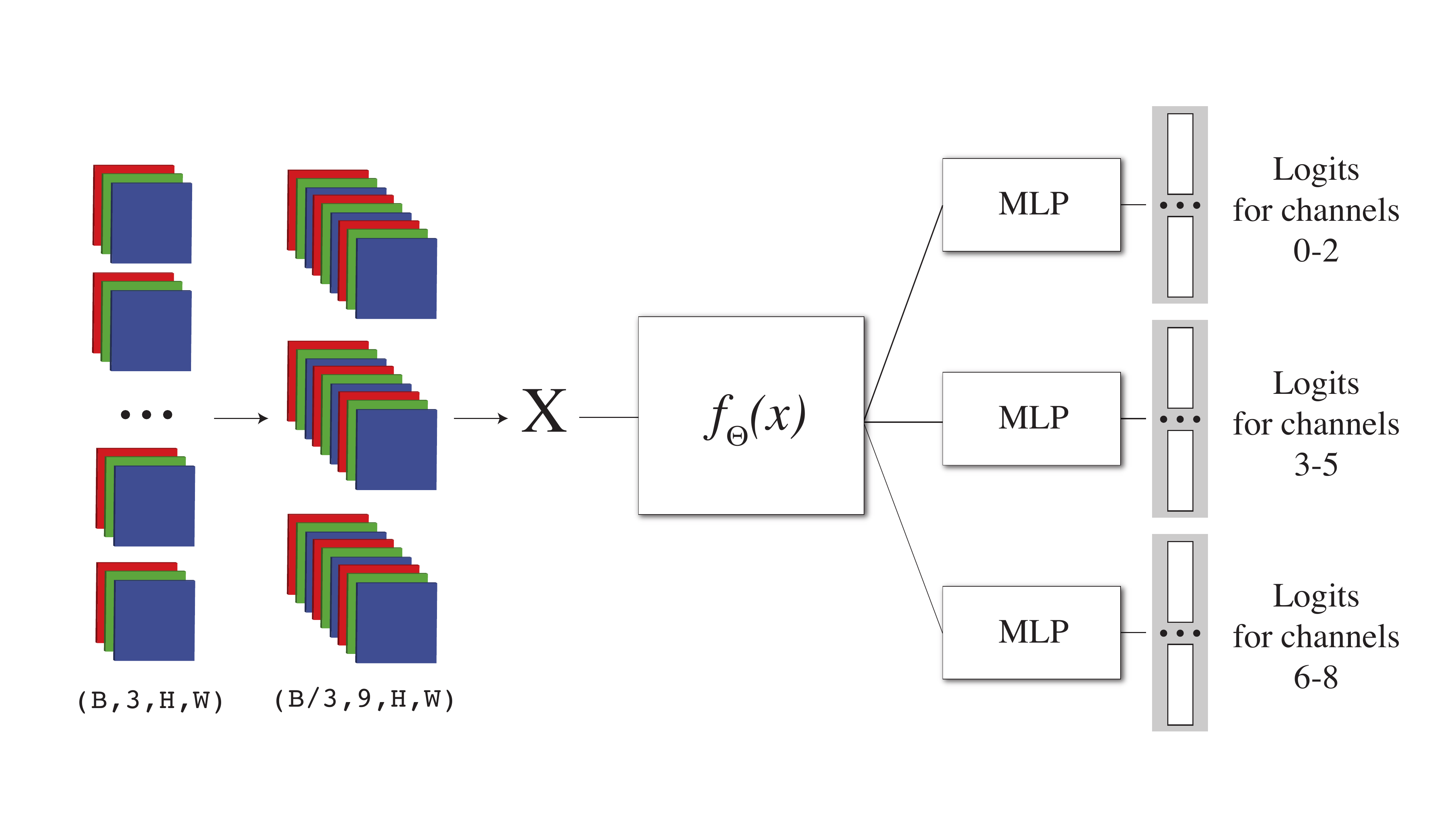}
\vspace{-8pt}
\caption{\textbf{High-level ImageNet Pretraining Network Setup.} A standard minibatch of shape $B \times 3 \times H \times W$ is reshaped to have a channel dimension of $3F = 3 \cdot 3 = 9$. This is passed into a grouped-convolutional feature extractor $f_\theta(x)$, and then passed to 3 MLP heads, each of which predict logits for a subset of the channels in $\textbf{I}$. Red, Green, and Blue squares represent the RGB channels in a standard image. Note that $\textbf{I}$ consists of the RGB channels for 3 images stacked together.
 }
\label{fig:imagenet_architecture1}
\end{figure}

\begin{figure}[t]
  \centering
    \includegraphics[width=.47\textwidth]{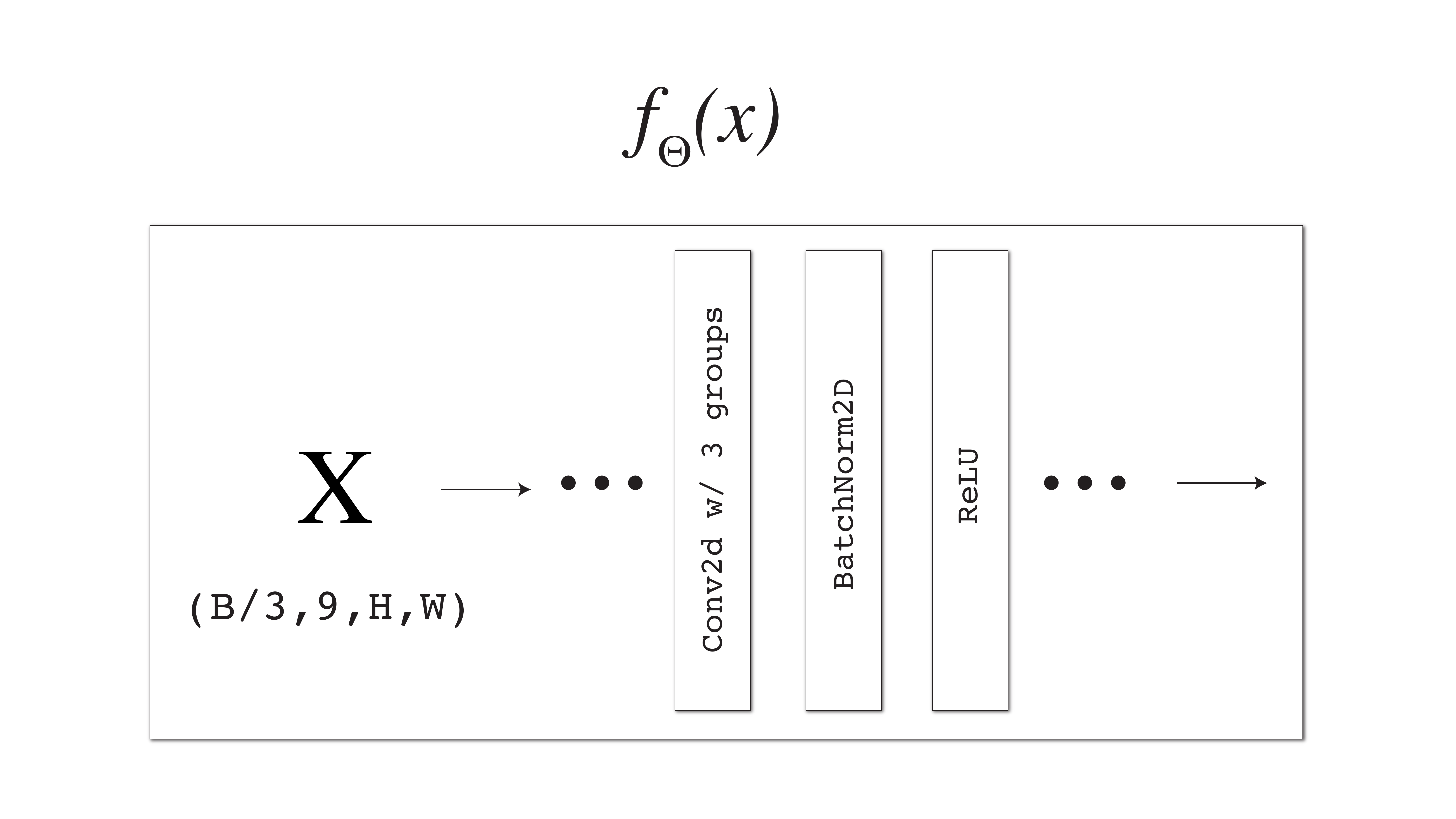} 
\vspace{-8pt}
\caption{\textbf{Grouped Convolutional Feature Extractor.} The feature extractor follows the classic Conv-BatchNorm-ReLU pattern. Since all standard convolutions are replaced with grouped convolutions with 3 groups, the feedforward signal from channels $0-2$, $3-5$, and $6-8$ are all independent. The first $\frac{1}{3}$ of the channels of the output correspond to channels $0-2$ of the input, and so on for channels $3-5$ and $6-8$ of the input. Thus, each MLP head in Figure~\ref{fig:imagenet_architecture1} is only passed the channels which correspond to the image they are trained to classify. This results in a fast network architecture where the feedforward signal for the inputs is never mixed.
 }
\label{fig:imagenet_architecture2}
\end{figure}

\label{subsubsec:imagenet-arc}
To motivate our network architecture, recall that in pixel-based RL applications, the goal is to learn a policy $\pi(a_t | o_t)$, where the choice of observation $o_t$ is usually not a single image. Often, several successive still images are combined in a \textit{frame stack} of size $F$ to form $o_t$. This is commonly done because many useful metrics such a velocity and acceleration cannot be computed from just a single still image. In particular, since each image is a tensor of shape $3 \times H \times W$, $o_t$ is a tensor of shape $3F \times H \times W$. In order to handle such inputs, convolutional feature extractors commonly used in pixel-based RL tasks often start with a 2D convolutional layer with $3F$ input channels. 

In contrast, inputs for the ImageNet supervised learning task are of shape $3 \times H \times W$. This presents a challenge for pretraining using the ImageNet task, since input shapes for the ImageNet task and our RL tasks are different. We work around this issue by slightly modifying the ImageNet task and altering the pretraining network architecture. Given a single input image and label pair $(\textbf{I}, y)$, the standard loss used for ImageNet training is the cross-entropy loss between the predicted and true class label. In our pretraining setup, a single datapoint consists of $F$ random samples from the ImageNet dataset: $\{ (\textbf{I}_1, y_1), (\textbf{I}_2, y_2) \ldots (\textbf{I}_F, y_F) \}$. We then feed all $F$ inputs $\{ \textbf{I}_1, \textbf{I}_2, \ldots \textbf{I}_f \}$ into our network at once, resulting in an input shape of $3F \times H \times W$. Notably, this input shape is identical to that which will be used during RL training. The network is tasked with classifying all $F$ inputs at once, so our pretraining setup uses the following loss for a single datapoint:

$$ \mathcal{L}(\theta) = \frac{1}{F} \sum_{i=0}^F \mathcal{L}_{CE}(y_i, f_{i, \theta}(\hat{y} | \textbf{I}_i)) $$

where $\mathcal{L}_{CE}$ denotes the cross-entropy loss between the predicted and true class label for the sample at index $i$. 

The feature extractor accepts this input of shape $3F \times H \times W$, and passes it through a series of Conv-BatchNorm-ReLU layers and then $F$ linear output heads, which classify each of the $F$ input images. During preliminary testing, we notice that this method of pretraining achieves very poor accuracy, even on a subset of 100 out of 1,000 ImageNet classes. We hypothesize that standard convolutions mix the feedforward signal of each image, making it increasingly difficult for the linear output heads at the end of the network to extract the correct relevant information from each image.

In order to solve this issue of ``signal mixing", our final network architecture consists of grouped convolutional layers. A grouped convolution with $N$ groups will take a set of $kN$ channels as input, and apply $N$ independent convolutions on channels $\{1, \ldots k\}, \{k + 1, \ldots 2k\}, \ldots \{(N-1)k + 1, \ldots , N k\}$. Given an input of size $3F \times H \times W$, we utilize a grouped convolution with $N = F$ groups in order to apply a different convolutional filter to each $3 \times H \times W$ input image in parallel.

In our experiments we select $F = 3$, following previous work on DeepMind Control tasks \citep{hansen2020selfsupervised}. Figure~\ref{fig:imagenet_architecture1} and Figure~\ref{fig:imagenet_architecture2} outline our network architecture in detail. Using this setup, we observe faster convergence and higher accuracy during ImageNet pretraining. For all RL experiments with ImageNet pretraining, we pretrain our fully parallel network for 400 epochs on 100 ImageNet classes. We select 100 random classes from those used in the ImageNet-R dataset \citep{hendrycks2020faces}, which consists of visually dissimilar classes. Since the networks used in RL training are not usually as powerful as those used to solve ImageNet, we do not want to spend inordinate amounts of time learning to separate closely related classes, such as ``Norwich terrier" and ``Norfolk terrier".

\subsection{Self-Supervised Pretraining: Inverse Kinematics (IK)}

Due to concern about the clear distributional gap between ImageNet and the DM Control environments, we explore an additional pretraining strategy based on learning the inverse kinematics of each individual environment. In this setting, our goal is to learn weights for a supervised learning task, which will then be used in RL training. We gather data directly from the RL environments. This avoids potential distributional mismatch problems, at the cost of requiring additional environment steps compared to ImageNet-based pretraining.

\subsubsection{Data Generation}
\label{subsec:datagen}
\begin{figure}[t]
  \centering
    \includegraphics[width=.49\textwidth]{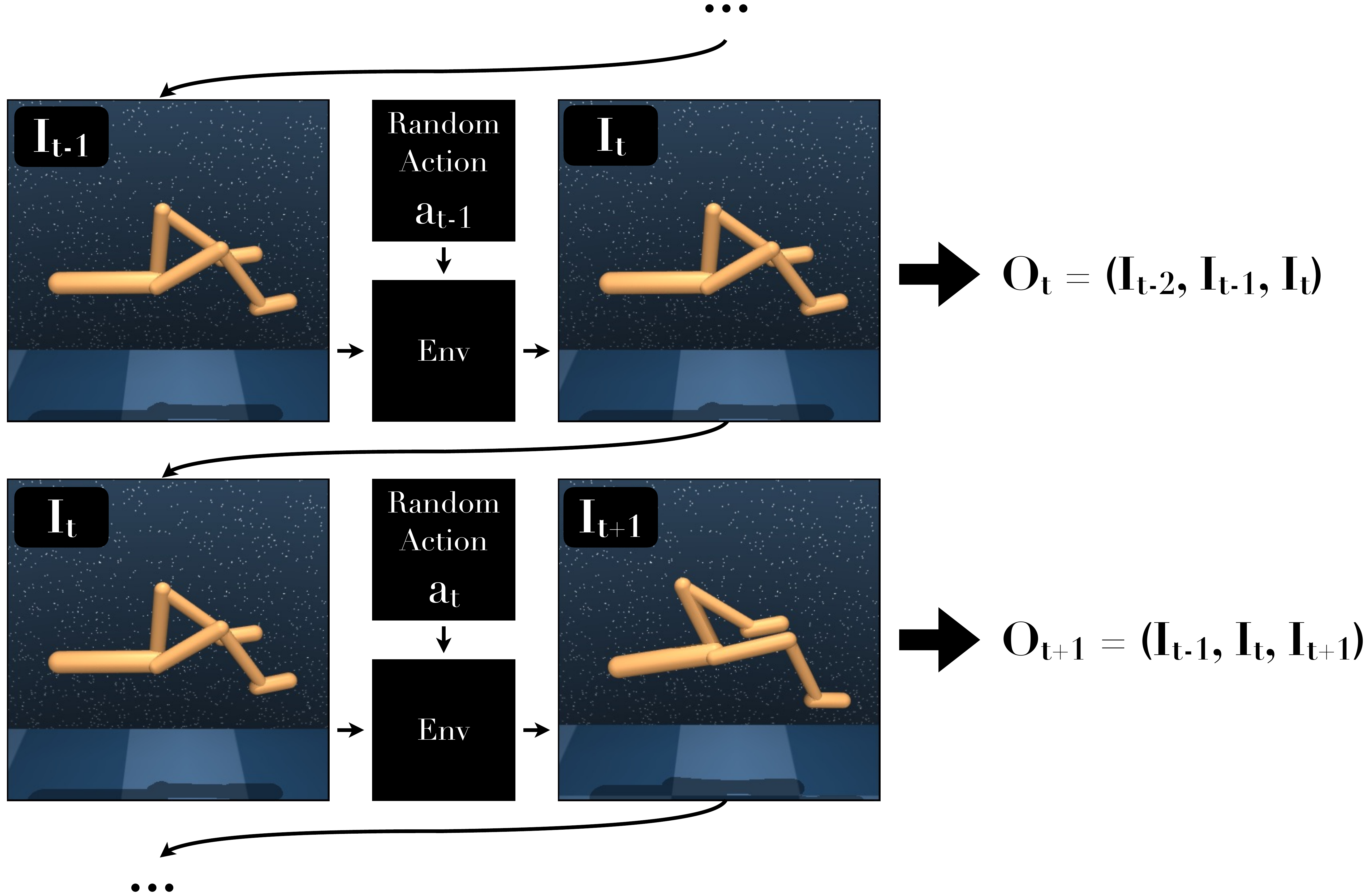}
\vspace*{-16pt}
\caption{\textbf{Self-Supervised Dataset Generation.} In order to conduct pretraining on the inverse kinematics of an environment, we collect triples $(o_t, o_{t+1}, a_t)$, where $o_{t+1}$ is the observation that results from taking random action $a_t$ from observation $o_t$. Repeating this process results in a self-supervised IK dataset appropriate for supervised learning. 
}
\label{fig:datagen}
\end{figure}

For each environment, we generate a dataset comprised of triples $(o_t, o_{t+1}, a_t)$. Here, $o_{t+1}$ is the observation that results from taking action $a_t$ from observation $o_t$. Each $o_t$ consists of three images. We collect these triples by initializing an environment and taking randomly sampled actions at every timestep, while resetting the environment every $k$ timesteps. We repeat this process until we obtain a dataset of size $T$, i.e. $\mathcal{D}_{\rm env} = \left\{(o_t, o_{t+1}, a_t)\right\}_{t=1}^{T}$. In particular, we take $T = 200000$ and $k = 100$. This data generation method is described visually in Figure~\ref{fig:datagen}. We apply the above procedure identically for both the normal and distracted environments.

\subsubsection{IK Training}
\label{subsec:ik-training}

Given $\mathcal{D}_{\rm env}$ for a specific environment, we learn the inverse kinematics of that environment. In particular, we frame this as a regression problem, with an input of ($o_t, o_{t+1}$), and output of $a_t$. Since the action spaces are continuous for each of the environments, we minimize the following mean-squared loss:

$$ \mathcal{L}(\theta) = \frac{1}{T} \sum_{t=1}^{T} \left\lVert a_t - f_\theta(o_t, o_{t+1})\right\rVert^2_2 $$\

\begin{figure}[t]
  \centering
    \includegraphics[width=.48\textwidth]{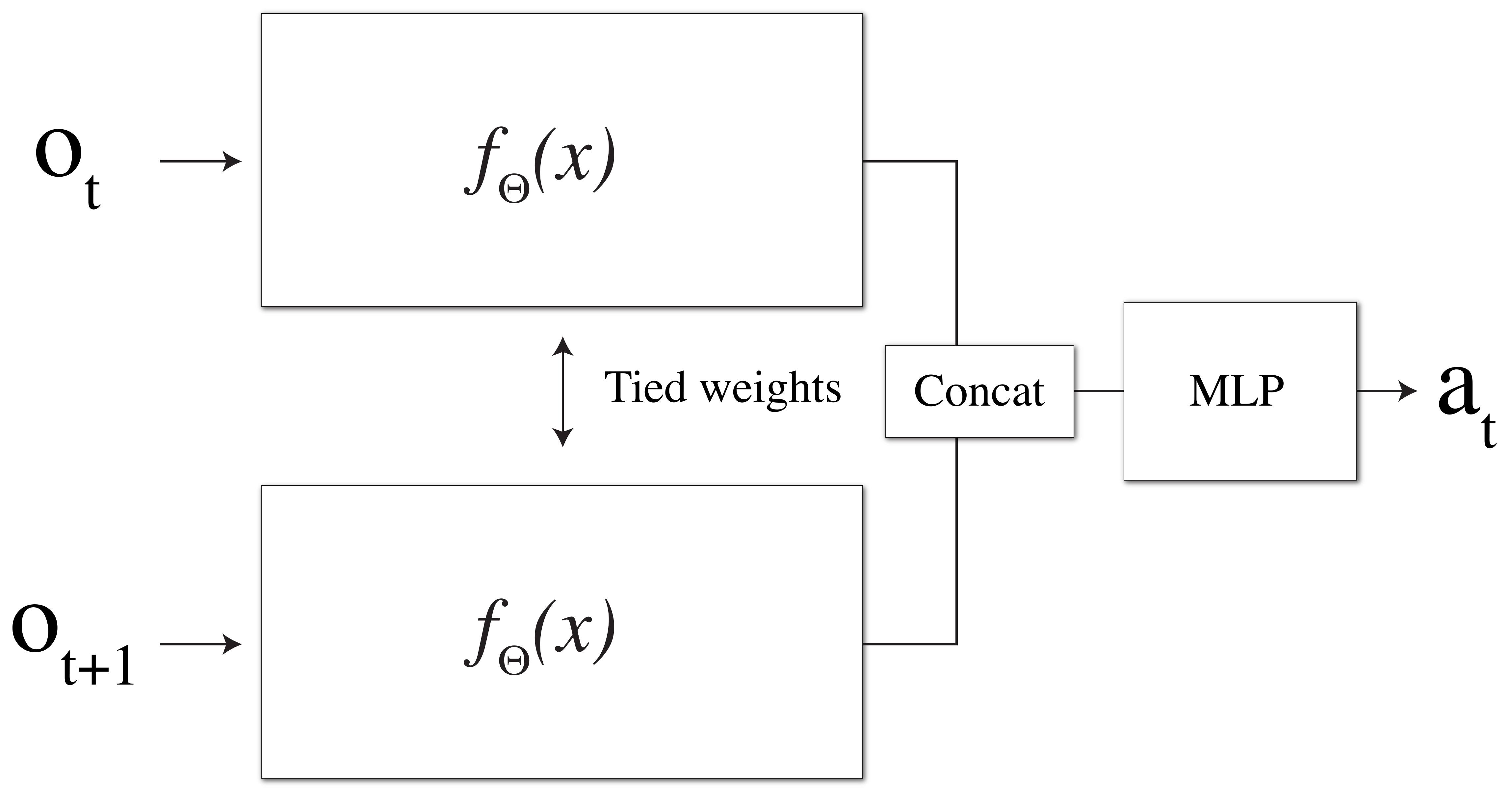}
\vspace*{-8pt}
\caption{\textbf{Inverse Kinematics Pretraining Setup.} Given a dataset comprised of $(o_t, o_{t+1}, a_t)$ triples (see Figure~\ref{fig:datagen}), IK pretraining is framed as a regression problem, with an input of ($o_t, o_{t+1}$) and an output of $a_t$. The weights of the convolutional layers are shared, and match the shape of the RL network. 
}
\label{fig:ik_setup}
\end{figure}

\begin{figure*}[t!]
\begin{subfigure}{.33\textwidth}
  \centering
    \includegraphics[width=\textwidth]{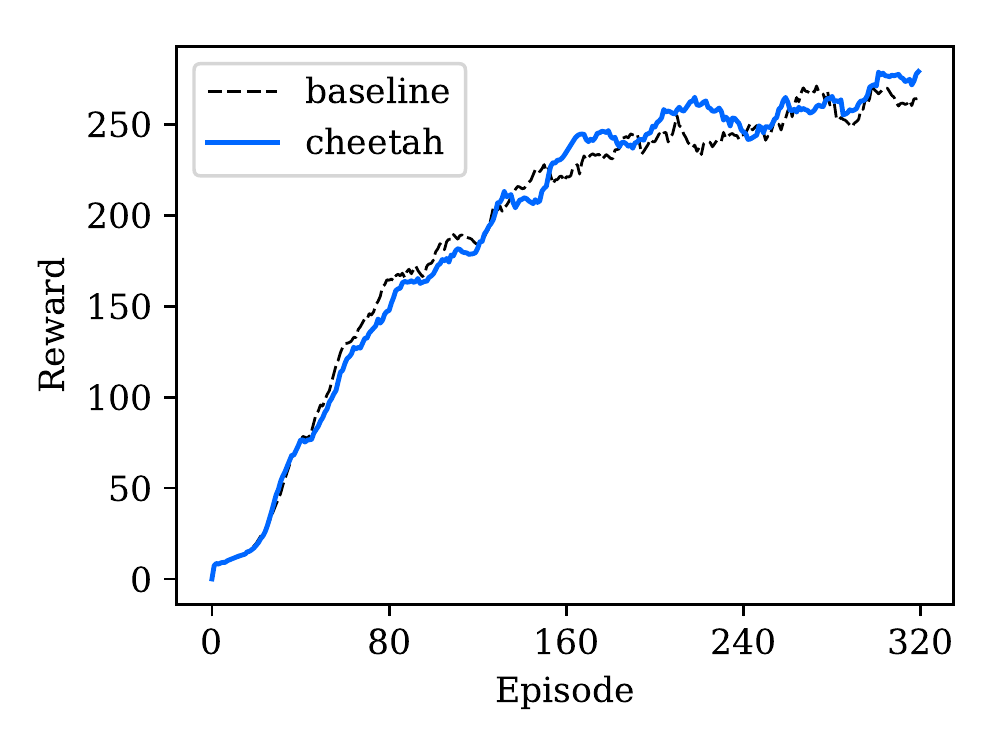}
    \vspace{-20pt}
  \caption{Cheetah}
  \label{fig:in_easy_a}
\end{subfigure}%
\begin{subfigure}{.33\textwidth}
  \centering
    \includegraphics[width=\textwidth]{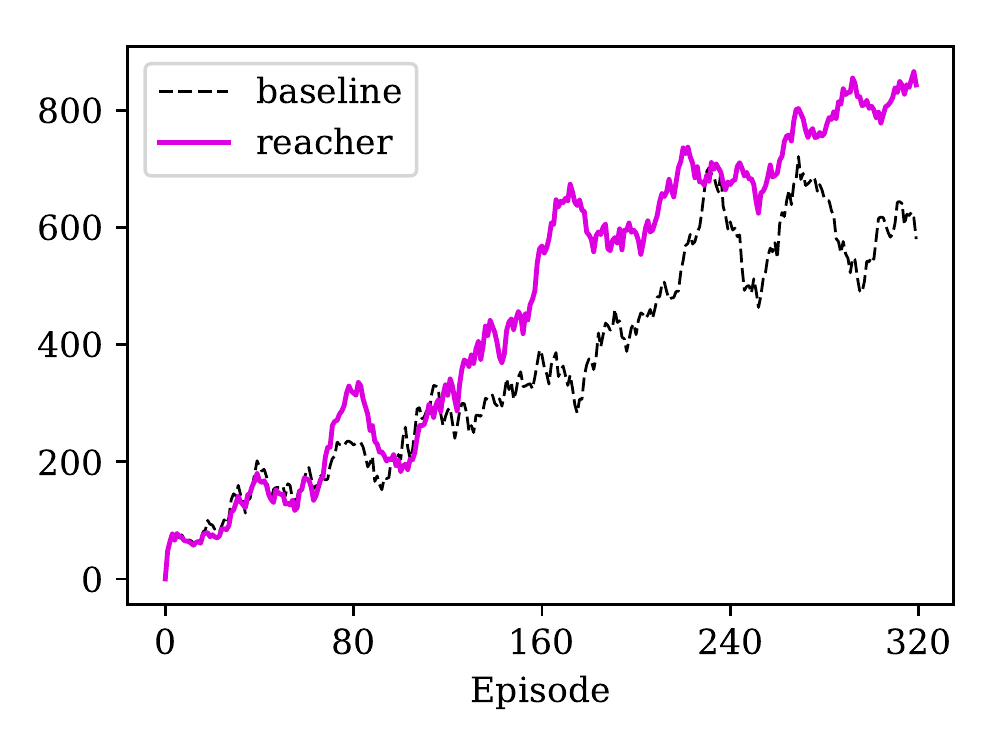}
    \vspace{-20pt}
  \caption{Reacher}
  \label{fig:in_easy_b}
\end{subfigure}%
\begin{subfigure}{.33\textwidth}
  \centering
    \includegraphics[width=\textwidth]{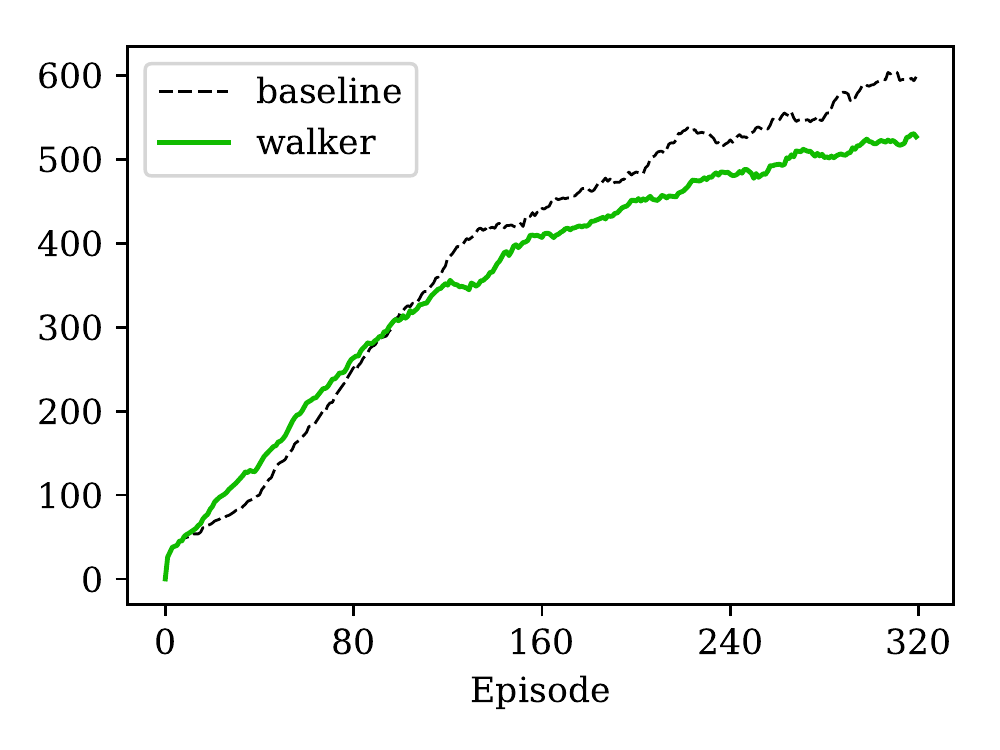}
    \vspace{-20pt}
  \caption{Walker}
  \label{fig:in_easy_c}
\end{subfigure}%
\caption{\textbf{Imagenet Pretraining Evaluation.} Each learning curve is the mean reward for each episode across 3 seeds, smoothed with a length-$15$ moving average. Pretraining is not particularly useful in terms of speed or performance for Cheetah and Walker, but helps substantially for Reacher. Curiously, the final score achieved by the Walker task is \textit{lower} with pretraining. The baseline uses random initialization. 
}
\label{fig:in_easy}
\end{figure*}

\begin{figure*}[t!]
\begin{subfigure}{.33\textwidth}
  \centering
    \includegraphics[width=\textwidth]{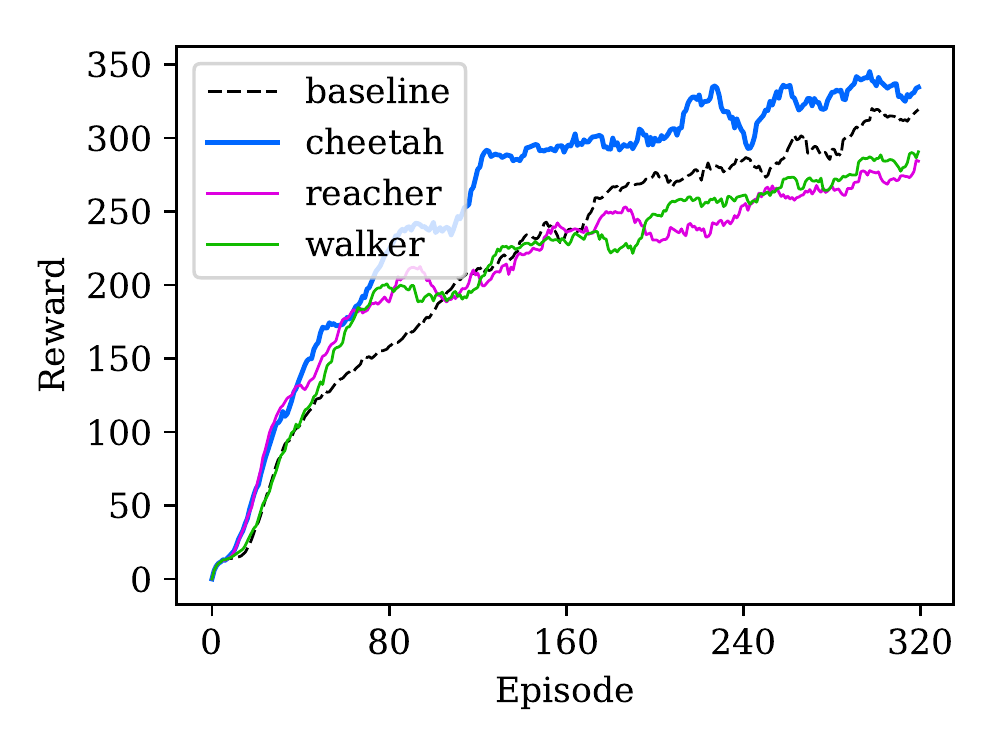}
    \vspace{-20pt}
  \caption{Cheetah}
  \label{fig:ik_easy_a}
\end{subfigure}%
\begin{subfigure}{.33\textwidth}
  \centering
    \includegraphics[width=\textwidth]{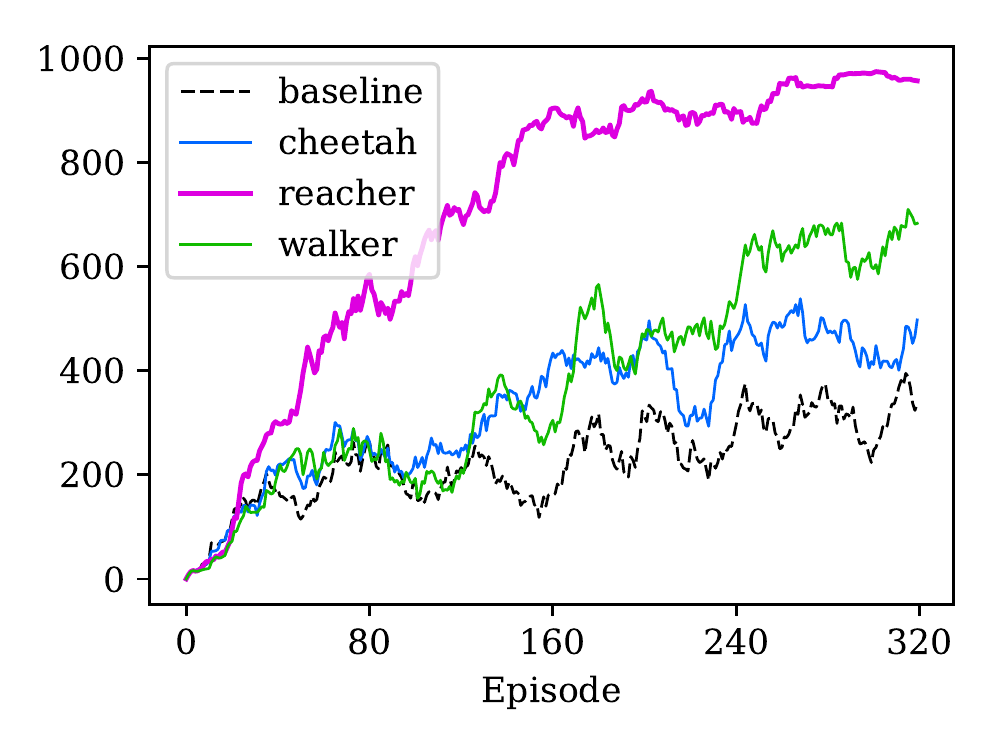}
    \vspace{-20pt}
  \caption{Reacher}
  \label{fig:ik_easy_b}
\end{subfigure}%
\begin{subfigure}{.33\textwidth}
  \centering
    \includegraphics[width=\textwidth]{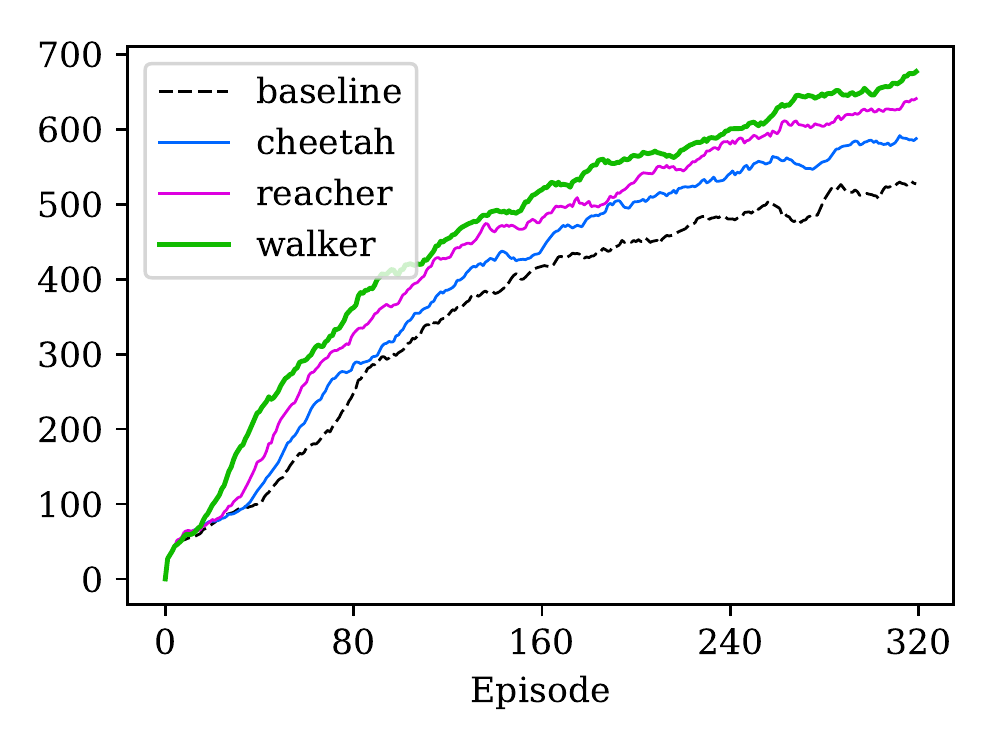}
    \vspace{-20pt}
  \caption{Walker}
  \label{fig:ik_easy_c}
\end{subfigure}%
\caption{\textbf{IK Pretraining Evaluation.} Each learning curve is the mean reward for each episode across 3 seeds, smoothed with a length-$15$ moving average. The caption details the RL training environment, and the curves are colored by the pretraining environment. Across all three environments, we observe significant improvements in performance over the baseline by using pretraining, with the largest benefit resulting from pretraining on the same RL environment. In the Reacher and Walker environment, pretraining helped substantially regardless of the IK training environment, while for Cheetah, the benefits are only present when pretraining on Cheetah. The baseline uses random initialization.  
}
\label{fig:ik_easy}
\end{figure*}

Figure~\ref{fig:ik_setup} describes our network architecture during IK pretraining. Since IK training requires the architecture to take both $o_t$ and $o_{t+1}$ as input, we use 2 convolutional feature extractors with tied weights. These feature extractors feed their concatenated hidden representations to a small MLP with Linear, Dropout, and ReLU layers, which outputs a predicted action $a_t$.

For each environment we train an IK model using the environment-specific datasets described in Section~\ref{subsec:datagen}. Then, for each of the three RL tasks, we perform four sets of Soft Actor-Critic experiments (each repeated over three seeds): first, a baseline case where we randomly initialize weights for both the Actor and the Critic; then, three separate cases where we use weights from each of the three pretrained IK models to initialize the two networks.

Although this method of pretraining requires access to the environment beforehand, it does not require any knowledge of the reward structure. This method of pretraining could be especially useful in complex pixel-base environments where the reward structure is difficult to evaluate, allowing the agent to learn feature extraction before having to use any reward samples during RL training.

\subsection{RL Training}


After pretraining, we load our feature extractor's weights into our RL training setup. Since the weights for all convolutional layers are shared between the Actor and the Critic, weights for both networks are loaded at once.

We then begin RL training as normal, using Soft Actor-Critic, as described in Section~\ref{subsec:sac}. The Actor takes the feature extractor's hidden representation and outputs a distribution over actions, $\pi(o_t)$. The Critic takes the same hidden representation and an action, and outputs a prediction for $Q(o_t, a_t)$. Further details for our RL agent architecture can be found in \cite{hansen2020selfsupervised}, as we built off of their implementation. In the case of our ImageNet pretraining experiments, however, this feature extractor consists fully of grouped convolutions and a larger hidden representation size, as described in Section~\ref{subsubsec:imagenet-arc}. Since grouped convolutions drastically reduce the number of model parameters, we used $3 \times$ the number of filters in the feature extractor for our ImageNet experiments than we did for our IK experiments. This results in different baselines for each set.

Since our main goal is to improve data efficiency and encourage quick model convergence, we run all experiments for 80k iterations. The standard amount of steps to run DeepMind Control environments ranges from 500k to 2000k steps. Our choice of 80k steps to evaluate data efficiency is based on previous work that uses 100k steps \citep{laskin2020reinforcement}. We were unable to match 100k steps due to time and memory limitations arising from our unbounded replay buffer size. 


\section{Experimental Results}


\subsection{Pretraining Results}

\subsubsection{ImageNet}

\begin{figure*}[t!]
\begin{subfigure}{.33\textwidth}
  \centering
    \includegraphics[width=\textwidth]{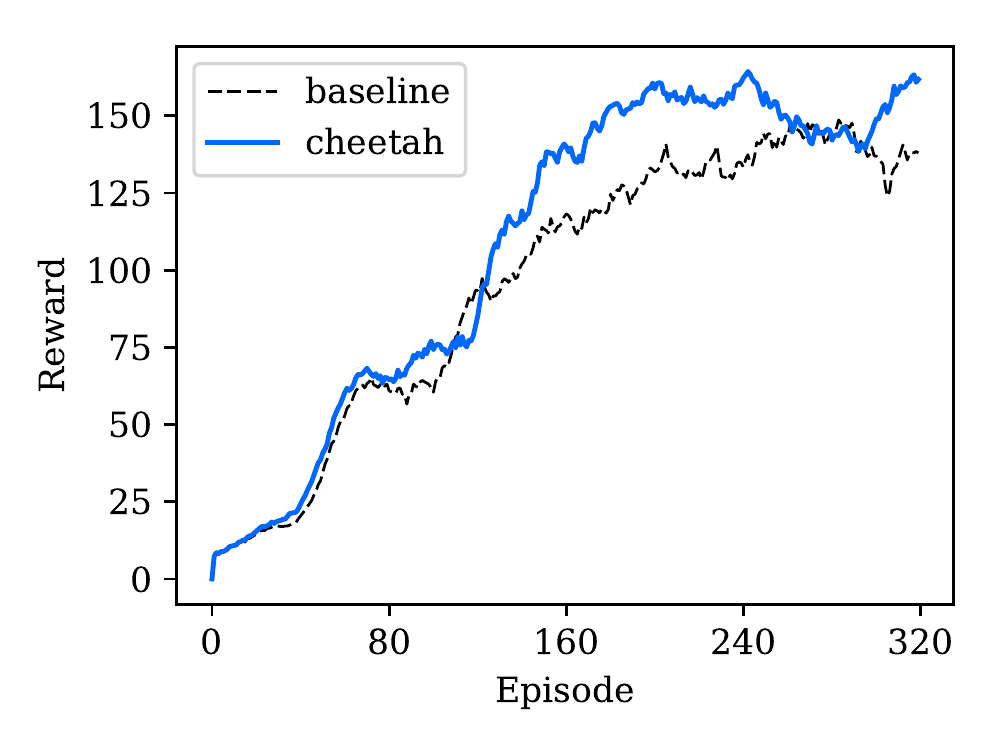}
    \vspace{-20pt}
  \caption{Cheetah}
  \label{fig:in_hard_a}
\end{subfigure}%
\begin{subfigure}{.33\textwidth}
  \centering
    \includegraphics[width=\textwidth]{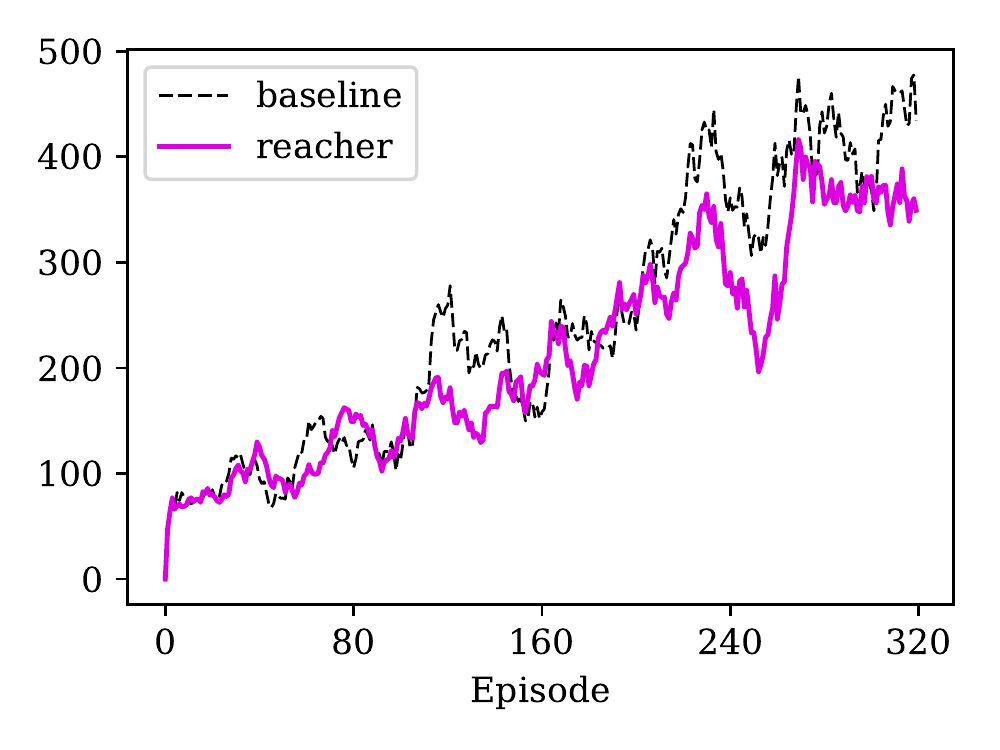}
    \vspace{-20pt}
  \caption{Reacher}
  \label{fig:in_hard_b}
\end{subfigure}%
\begin{subfigure}{.33\textwidth}
  \centering
    \includegraphics[width=\textwidth]{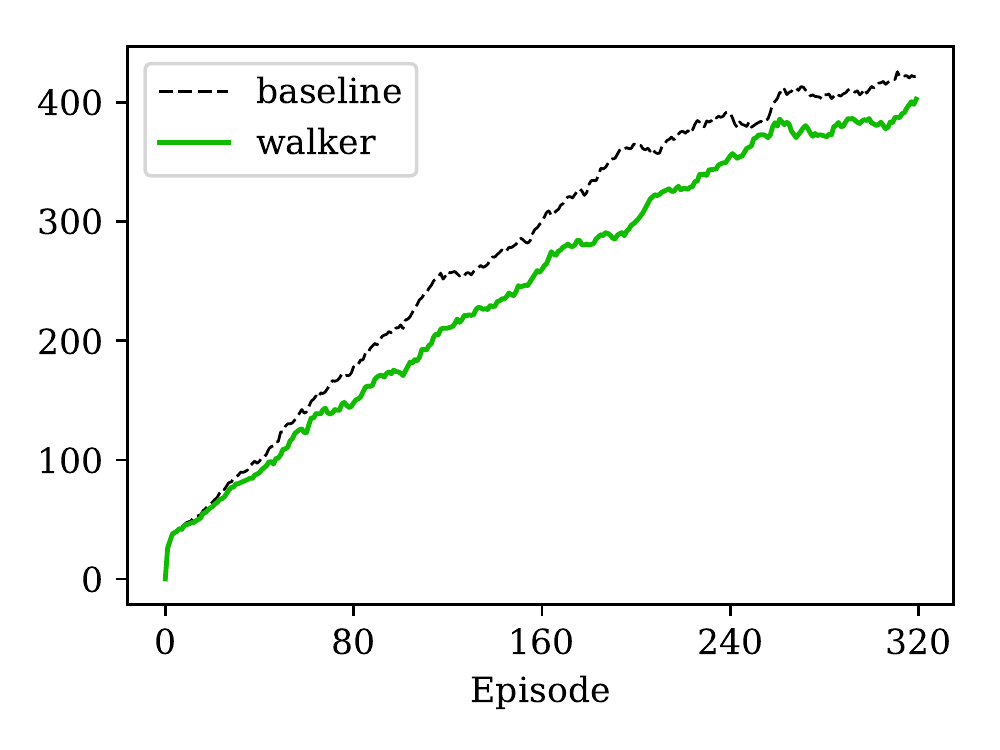}
    \vspace{-20pt}
  \caption{Walker}
  \label{fig:in_hard_c}
\end{subfigure}%
\caption{\textbf{ImageNet Pretraining with Distracting Background Evaluation.} Each learning curve is the mean reward for each episode across 3 background videos, smoothed with a length-$15$ moving average. On the Reacher and Walker environments, pretraining does not provide any significant improvement, while on the Cheetah environment, it provides a slight improvement in performance. The baseline uses random initialization. 
}
\label{fig:in_hard}
\end{figure*}

\begin{figure*}[t!]
\begin{subfigure}{.33\textwidth}
  \centering
    \includegraphics[width=\textwidth]{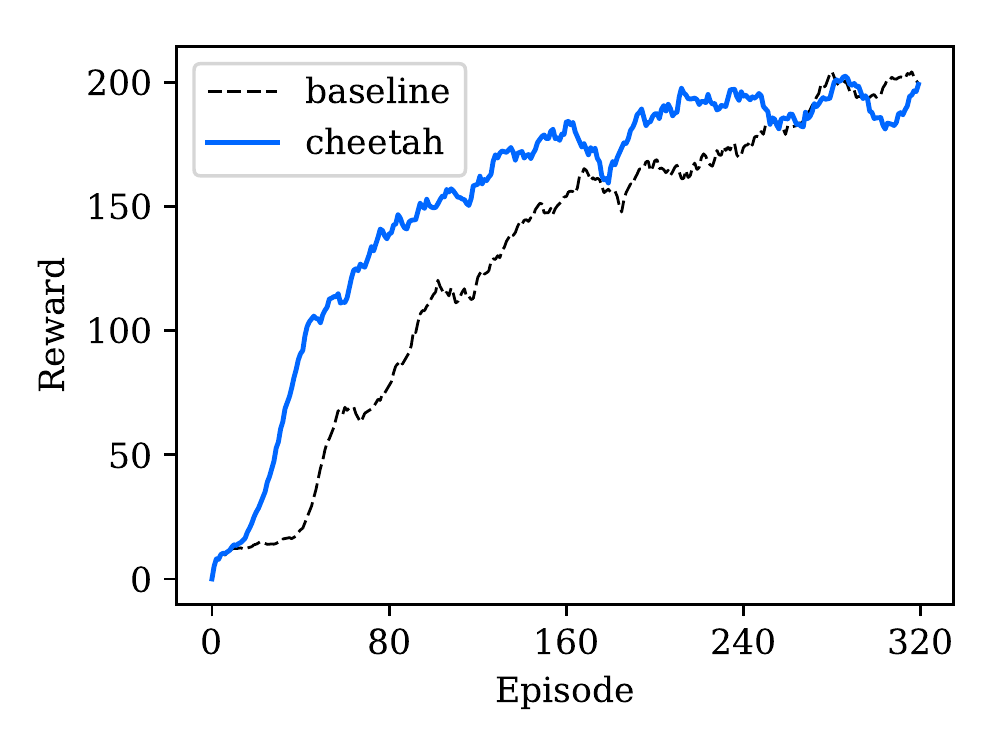}
    \vspace{-20pt}
  \caption{Cheetah}
  \label{fig:ik_hard_a}
\end{subfigure}%
\begin{subfigure}{.33\textwidth}
  \centering
    \includegraphics[width=\textwidth]{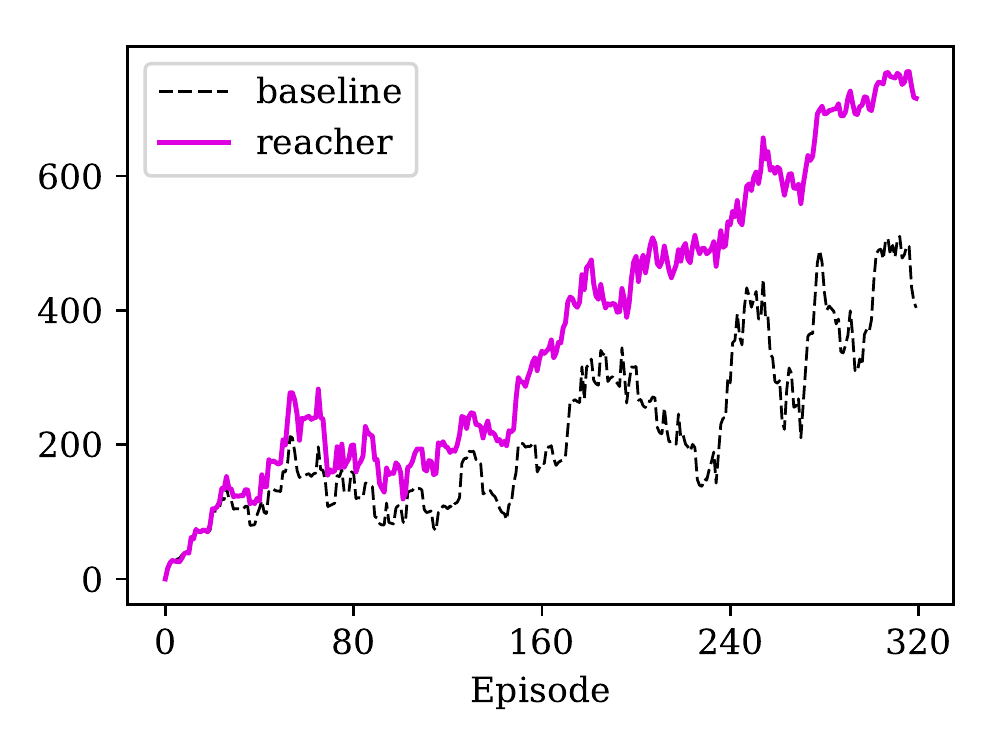}
    \vspace{-20pt}
  \caption{Reacher}
  \label{fig:ik_hard_b}
\end{subfigure}%
\begin{subfigure}{.33\textwidth}
  \centering
    \includegraphics[width=\textwidth]{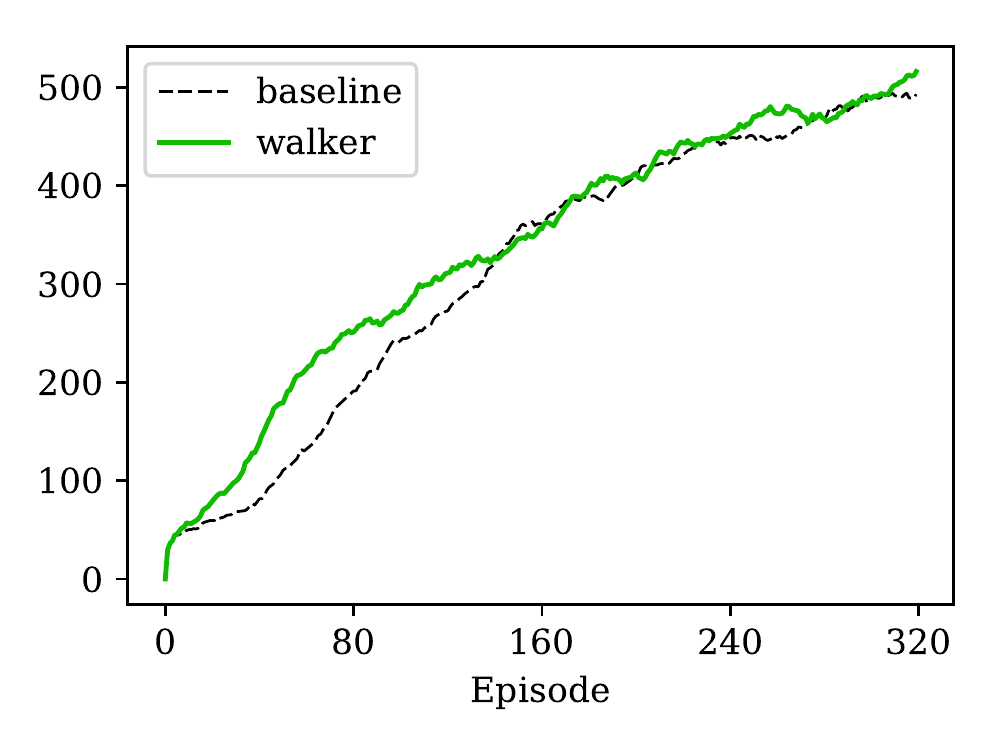}
    \vspace{-20pt}
  \caption{Walker}
  \label{fig:ik_hard_c}
\end{subfigure}%
\caption{\textbf{IK Pretraining with Distracting Background Evaluation.} Each learning curve is the mean reward for each episode across 3 background videos, smoothed with a length-$15$ moving average. IK-based pretraining provides a significant improvement to RL training in the distracted Reacher environment, though in the Walker and Cheetah environment, despite early advantages, both the baseline and pretrained network ultimately appear to converge to a similar value. The baseline uses random initialization. 
}
\label{fig:ik_hard}
\end{figure*}

Results are shown in Figure~\ref{fig:in_easy}. Each curve represents the mean reward over 3 seeds for each experiment. Pretraining does not help on the Walker or Cheetah environments, but it does seem to help on the Reacher environment, finishing over 200 points higher than the baseline across all 3 seeds. Curiously, the final score achieved by the Walker task is \textit{lower} with pretraining. Although the final score achieved is lower, higher scores are achieved earlier during training, between episodes 0 and 100. Regardless, the effectiveness of ImageNet pretraining is limited on the DeepMind Control suite, perhaps mainly because of the distribution mismatch between ImageNet and RL observations. 

\begin{figure*}[t]
\begin{subfigure}{.33\textwidth}
  \centering
    \includegraphics[width=\textwidth]{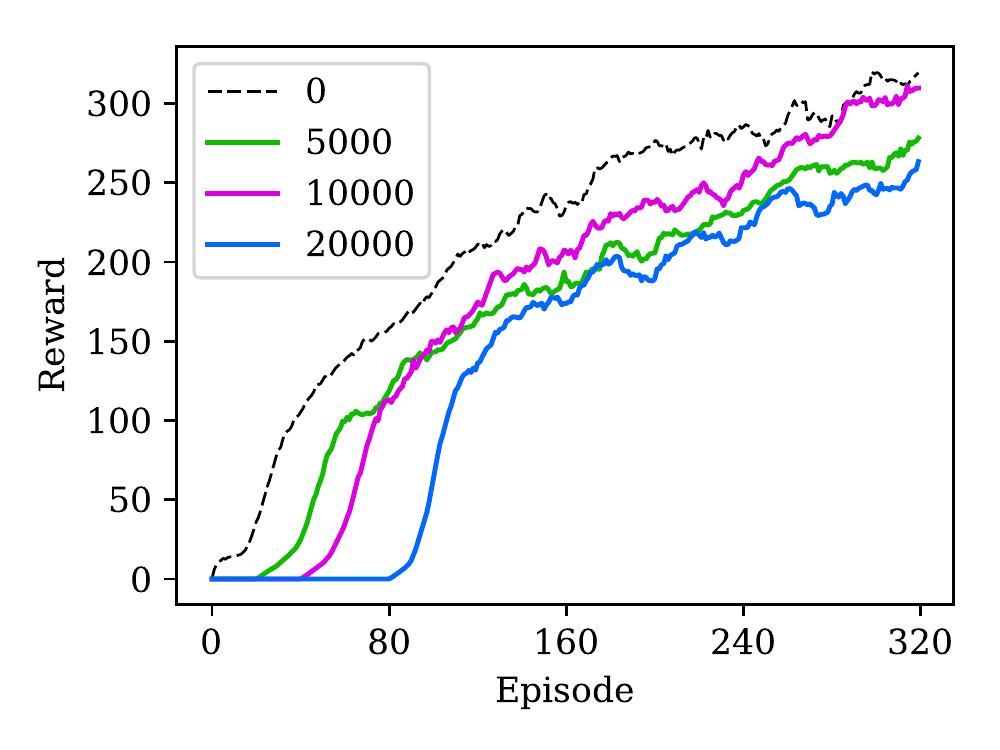}
    \vspace{-20pt}
  \caption{Cheetah}
  \label{fig:ik_hard_a}
\end{subfigure}%
\begin{subfigure}{.33\textwidth}
  \centering
    \includegraphics[width=\textwidth]{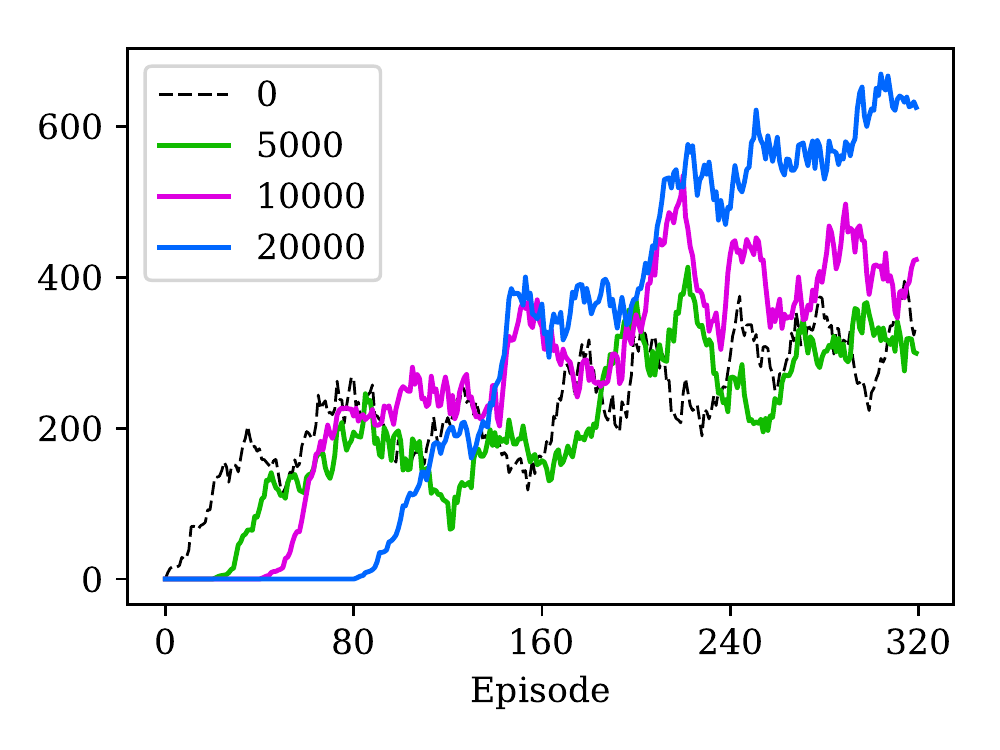}
    \vspace{-20pt}
  \caption{Reacher}
  \label{fig:ik_hard_b}
\end{subfigure}%
\begin{subfigure}{.33\textwidth}
  \centering
    \includegraphics[width=\textwidth]{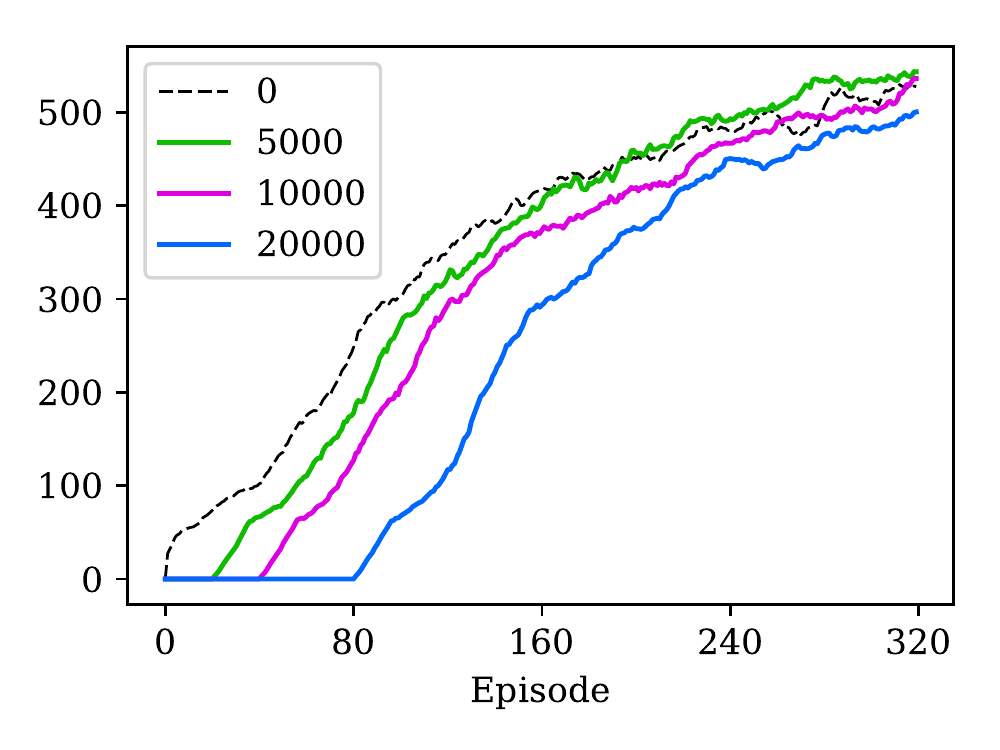}
    \vspace{-20pt}
  \caption{Walker}
  \label{fig:ik_hard_c}
\end{subfigure}%
\caption{\textbf{Sharpening the Axe.} Each learning curve is the mean reward for each episode across 3 seeds, smoothed with a length-$15$ moving average. The caption details the RL training environment, and the color of the curve corresponds to pretraining dataset size ($T\in\{0,5000,10000,20000\}$). Results suggest the benefits of inverse kinematic pretraining are limited when the number of environment steps allowed is small. The baseline, $T=0$, corresponds to random initialization. 
}
\label{fig:ik_axe}
\end{figure*}

\subsubsection{Inverse Kinematics}

Learning curves for the experiments described in  Section~\ref{subsec:ik-training} are presented in Figure~\ref{fig:ik_easy}. Each curve represents the mean reward over 3 seeds. Across all three environments, we observe significant improvements in performance over the baseline by using pretraining. The improvement is especially dramatic in the Reacher environment, though more modest gains can also be observed in both the Cheetah and Walker environments.

Furthermore, as one may expect, pretraining is more effective in the intra-environment case: that is, when the inverse kinematics model is trained on data from the same environment as where RL training is performed, the benefit provided by pretraining is more significant. This is likely due to a better match in distributions between the pretraining and RL training phases. Once more, this effect is most noticeable for Reacher, but the intra-environment pretraining experiments still exhibit the highest performance for both Cheetah and Walker. In fact, we actually observe \textit{worse} performance compared to the baseline for the Cheetah environment in the cross-environment cases where we pretrain on Reacher or Walker data.

Overall, we find that our self-supervised strategy for pretraining does provide improved performance on the DeepMind Control Suite. In particular, this pretraining method provides a much clearer benefit compared to pretraining on ImageNet, which is perhaps due to reduced issues with distributional mismatch.

\subsection{Pretraining Results with Distracting Environments}

\subsubsection{ImageNet}

Results for ImageNet pretraining with distracting environments are in Figure~\ref{fig:in_hard}. Each learning curve is the average over running on 3 different distracting backgrounds. On the Reacher and Walker environments, we don't observe any significant improvement. On the Cheetah environment, there is a slight improvement in performance. 

Although the backgrounds used for the distracting environments were somewhat closer in distribution to ImageNet images, this did not lend much in terms of final performance. A possible reason for this is that the ImageNet pretraining setup forces the network to learn image features that may not necessarily be useful to compare images across the frame stack. In contrast, our inverse kinematics pretraining setup does not have this limitation.

\subsubsection{Inverse Kinematics}
We also applied our inverse kinematics pretraining method to RL in environments with distracting backgrounds. Results for these experiments are presented in Figure~\ref{fig:ik_hard}. In contrast to our initial IK pretraining, we do not conduct cross-environment evaluation. That is, for a given environment, we perform two sets of experiments, both over three distinct background videos: baseline experiments with random initialization, and experiments where we pretrain on IK data generated from the same environment, also with a distracting background.

Similarly to the normal case, IK-based pretraining provides a significant improvement to RL training in the distracted Reacher environment. The pretrained learning curve for Cheetah also initially outpaces the baseline by a sizable margin, though they ultimately appear to converge to a similar value. However, performance is very similar in the Walker environment besides a small early gain, though this is still an improvement on the performance of our ImageNet-based pretraining in this distraction environment.

\subsection{Maximizing Data Efficiency with IK Pretraining}

After observing improvements in RL performance via IK pretraining for both normal and distracted environments, we next undertake a somewhat more precise investigation of whether pretraining provides a net efficiency improvement. In particular, given a limited ``budget" of environment steps, two natural questions arise: (1) whether one should allocate some of those environment steps towards obtaining IK pretraining data, and (2) if so, exactly how many steps to allocate, before moving onto the RL training phase.

To address these questions, we let $N$ represent the maximal number of agent steps one is allowed to take in an environment (i.e. the budget). In practice, $N$ may be limited due to cost or time constraints, so deciding how best to utilize data from the environment may be a useful optimization. To this end, we let the hyperparameter $T < N$ represent the number of agent steps one spends collecting a dataset for IK pretraining. For example, $T = 0$ corresponds to performing no pretraining prior to RL training. Note that in our initial experiments, we took $T = 200000$. 

We fix $N = 80000$ and evaluate performance for $T \in \{0,5000,10000,20000\}$ across 3 seeds for each of the 3 environments. For a given environment and $T$, we first collect $T$ $(o_t, o_{t+1}, a_t)$ triples from the environment, and perform IK training via the same procedure described in Section~\ref{subsec:ik-training}. Then, we initialize RL training using these weights, and continue RL training for $N - T$ timesteps. Thus, for all values of $T$, we take exactly $N$ environment steps across the IK and RL training phases.

Results are inconclusive for $N = 80000$, as not enough steps are available for pretraining. The results for different values of $T$ on each environment are shown in Figure~\ref{fig:ik_axe}. Results suggest the benefits of inverse kinematic pretraining are limited when the number of environment steps allowed is small. The comparative ineffectiveness of pretraining with $T \ll 200000$ suggests the pretrained IK network struggles to learn from fewer samples. To quantify this, for all values of $T$, the IK training ran for 30 epochs. Thus, for $T=20000$, training consisted of only $\frac{1}{10}$ of the gradient updates compared to when $T=200000$. One possible way to mitigate this issue is to increase the number of epochs in proportion to the amount of data. Since the data is created by taking random actions over several episodes, the distribution of trajectories and observations in the pretraining dataset is likely not diverse. That is, it is unlikely that Walker randomly stood up and started walking-it is more likely that it spends the entirety of its time flopping around the ground aimlessly. We hypothesize that the lack of diversity should mean that smaller $T$ should still give performance gains that are comparable to $T=200000$. However, is unclear if increasing training epochs would cause pretraining to regain effectiveness or if it would simply overfit. We leave this analysis to future work. As a note, many RL tasks in this domain are trained for 500k to 2000k steps, which would allow for larger $T$ while still maintaining a reasonable amount of RL training.

\newpage

\section{Discussion \& Future Work}

This paper addresses a general approach for pretraining for RL tasks. Results suggest pretraining with a self-supervised network on the inverse kinematics of a similar environment reliably reduces RL training time and improves performance. Results also suggest pretraining is not as useful in the case the total number of environmental steps allowed is small.   

Given the time and compute limitations we had for this project, there are several avenues we are excited to explore in the future. In terms of the data generation phase, perhaps taking random actions is naive, and a more representative dataset could be generated by using existing exploration policies during the self-supervision phase. The caveat here is requiring knowledge of the reward structure. Further, for the optimal use of limited environmental data, increasing the horizon would give much more conclusive results. Given our limitations, this horizon is capped at 80k steps. However, many RL tasks in this domain are trained for 500k to 2000k steps, which would provide a more clear answer for selecting the optimal number of pretraining steps.



\newpage

\balance
\bibliography{refs}
\bibliographystyle{icml2019}

\end{document}